\documentclass[letterpaper,10pt,conference]{ieeeconf}  % Comment this line out if you need a4paper

\IEEEoverridecommandlockouts                              % This command is only needed if 
\usepackage[colorlinks,linkcolor=black]{hyperref} % for links
\usepackage{graphicx} % for embedding graphics
\usepackage{booktabs} % for pretty tables
\usepackage{float}
\usepackage[table,xcdraw]{xcolor}
\usepackage[english]{babel}
\usepackage{hyperref}
\usepackage{amsmath}

\usepackage{amsthm}
\usepackage{cleveref}
\usepackage{array}
\usepackage{longtable}
\usepackage{float} 
\usepackage{multirow}
\usepackage{dblfloatfix}
\usepackage{array}
\usepackage{titlesec}
\usepackage{algpseudocode} % http://ctan.org/pkg/algorithmicx
\usepackage{algorithm} % http://ctan.org/pkg/algorithm
\usepackage{balance}

\theoremstyle{definition}
\newtheorem{definition}{Definition}[section]

\usepackage[backend=biber]{biblatex}
\IEEEtriggeratref{16}

\title{\LARGE \bf
 Automated Planning Domain Inference for Task and Motion Planning
}

\author{
Jinbang Huang$^{1,2}$,
Allen Tao$^{2}$,
Rozilyn Marco$^{1}$, 
Miroslav Bogdanovic$^{2}$, 
Jonathan Kelly$^{1}$, and
Florian Shkurti$^{2}$
\thanks{Corresponding author 
    {\texttt\small jinbang.huang@robotics.utias.utoronto.ca}
}
\thanks{$^{1}$Space and Terrestrial Autonomous Systems Lab, $^{2}$Robot Vision and Learning Lab, University of Toronto Robotics Institute.}
}

\begin{document}

\maketitle
\thispagestyle{empty}

%%%%%%%%%%%%%%%%%%%%%%%%%%%%%%%%%%%%%%%%%%%%%%%%%%%%%%%%%%%%%%%%%%%%%%%%%%%%%%%%
\begin{abstract}

Task and motion planning (TAMP) frameworks address long and complex planning problems by integrating high-level task planners with low-level motion planners. However, existing TAMP methods rely heavily on the manual design of planning domains that specify the preconditions and postconditions of all high-level actions.
This paper proposes a method to automate planning domain inference from a handful of test-time trajectory demonstrations, reducing the reliance on human design. Our approach incorporates a deep learning-based estimator that predicts the appropriate components of a domain for a new task and a search algorithm that refines this prediction, reducing the size and ensuring the utility of the inferred domain. Our method can generate new domains from minimal test time demonstrations, enabling robots to handle complex tasks more efficiently.
We demonstrate that our approach outperforms behaviour cloning baselines, which directly imitate planner behaviour, in terms of planning performance and generalization across a variety of tasks. Additionally, our method reduces computational costs and data amount requirements at test time for inferring new planning domains. 
\end{abstract}

%%%%%%%%%%%%%%%%%%%%%%%%%%%%%%%%%%%%%%%%%%%%%%%%%%%%%%%%%%%%%%%%%%%%%%%%%%%%%%%%
\section{Introduction}

Robot autonomy that generalizes to diverse environments requires efficient integration of complex real-time perception, decision-making, and motion planning. Existing motion planning algorithms reach their limits in complex and high-dimensional environments~\cite{Gonzalez2016-jl, Mohanan2018-ho, Paden2016-fa, Schwartz1988-ry, Zhou2022-vg} that are frequently encountered in modern robot applications~\cite{Liu2022-wj}. Task and motion planning (TAMP) frameworks address this challenge by employing high-level task planners to discretize a complex planning task into a sequence of manageable sub-tasks that low-level motion planners can complete.

While existing TAMP frameworks have solved many complex and long-horizon tasks in robotics~\cite{Garrett2021-ka, Lozano-Perez1979-zv}, they rely heavily on human-designed planning domains, which are sets of logical rules and constraints that provide a basis for high-level task plans.
%regulating how to generate task plans. 
Planning domains are restricted to predefined decision spaces and cannot easily be adapted to new tasks. This limitation drives the need for automated solutions that generate planning domains by leveraging information about existing, similar domains.
% ones as a foundation.
We aim to develop a method that learns the relationships between tasks and their respective planning domains from training data, enabling the automatic generation of a new planning domain based on just one or a few demonstrations from humans. 

This paper introduces an automated planning domain inference method that generates planning domains for new TAMP problems in a one-shot or few-shot manner, using demonstrations of continuous state-action trajectories. Our approach uses deep learning to predict domains for new tasks. To further improve feasibility and planning efficiency, we combine these predictions with search to reduce the size of the planning domains, making them as small as possible while ensuring that they are effective and reliable.

The primary contribution of this paper is (1) a graph attention network (GAT) trained on a small set of demonstrations to efficiently predict the planning domain for new tasks; (2) a guided generate-and-test search algorithm to refine the predicted planning domain, ensuring that planning is both feasible and efficient; and (3) a framework that is deployable on a real robot that integrates GAT prediction and search to infer the planning domain for new tasks, requiring only a single human demonstration. We evaluate our method and compare it to existing baselines in 12 different environments. Our method significantly improves planning performance, generalizability, and computational costs.

% \jinbang{
%     \begin{enumerate}
%         \item Faster search using learned prior from previous problems
%         \item Reduced number of demonstrations needed
%         \item Real: One human/manual demonstration of the task -- find domain, plan, execute using robot
%         \item Accounting for uncertain evaluation in domain search
%     \end{enumerate}

% }

\begin{figure}
\centering
\includegraphics[width=1\linewidth]{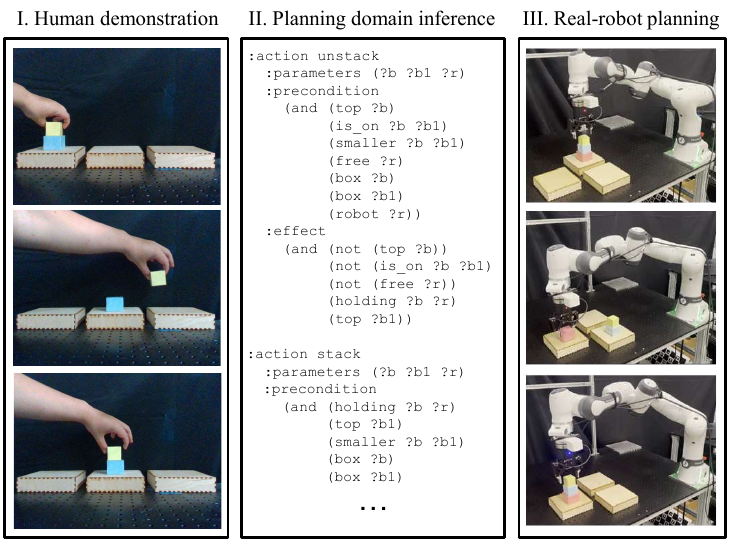}
\vspace{-6mm}
\caption{Illustration of the general idea of our method: (1) One or a few human demonstrations for task execution are provided. 3D perception is used to extract object poses and interactions. (2) Plausible planning domains are inferred based on human demonstrations. (3) The generated domain is used for task and motion planning on a real robot.}
\label{fig:Framework}
\vspace{-5mm}
\end{figure}

\section{Related Work}

\subsection{Learning to Plan for Faster Task and Motion Planning}

Recent advances in machine learning have brought new approaches to TAMP. Learning-based TAMP methods often train policies that either accelerate or replace traditional planners. Recent research highlights the effectiveness of both supervised learning and reinforcement learning. Supervised learning methods significantly accelerate planning, but they often require extensive demonstration data and are conditioned to specific environments~\cite{ Khodeir2023-so,Kim2020-du, Lin2022-tc, Silver2021-mv}. This limits their ability to generalize across different environment settings or goals~\cite{Dalal2023-cq, McDonald2021-uh, Zhu2021-sx, Yang2022-xa}. Reinforcement learning, as another major learning-based TAMP method in recent years, assists in accelerating planning speed and enhances adaptability to dynamic and uncertain environments~\cite{Chitnis2016-pd, Jiang2019-gy, Paxton2017-lj, Xu2021-md}. However, creating a well-defined learning environment for complex TAMP problems, in simulation or in the real world, can be challenging. This reduces the applicability of reinforcement learning in many TAMP scenarios~\cite{Wang2022-dx, Zhou2022-vg}. 

\subsection{Learning Planning Domains}

Recognizing the discussed limitations, researchers have increasingly shifted their focus toward learning planning domains. The mainstream learning-based methods in TAMP, designed to replicate the planner's behaviour, implicitly learn both the planning domain and the search algorithm. Focusing solely on learning the planning domain and integrating it with existing search algorithms reduces the learning burden. 

Success in learning logical predicates from the environment ~\cite{Kase2020-ef, Mukherjee2021-bf, Silver2023-mi} facilitates the study of automatic generation of planning domains~\cite{Kumar2023-nk}. Automatic planning domain generation relies on pre-existing logical predicates and actions to create planning domains autonomously. Recent studies have shown that learning a planning domain improves generalizability and adaptability compared to directly learning a TAMP planner~\cite{Diehl2021-qg}. However, existing methods encounter the challenges of high search costs and the requirement of a relatively large dataset~\cite{Kumar2023-nk, Silver2023-mi}.

Our method is closely related to that of Kumar et al. \cite{Kumar2023-nk}, which identifies necessary preconditions and postconditions of an operator through a hill-climbing search. However, we attempt to generate a full planning domain instead of merely operators (actions). Furthermore, rather than starting the search from scratch, our approach leverages transferable knowledge from existing planning domains, significantly accelerating the generation of new domains and reducing the number of demonstrations required during execution. As a result, we are able to generate new planning domains for unseen tasks in a one-shot or few-shot manner.

\begin{figure*}[t]
\centering
\includegraphics[width=0.976\linewidth]{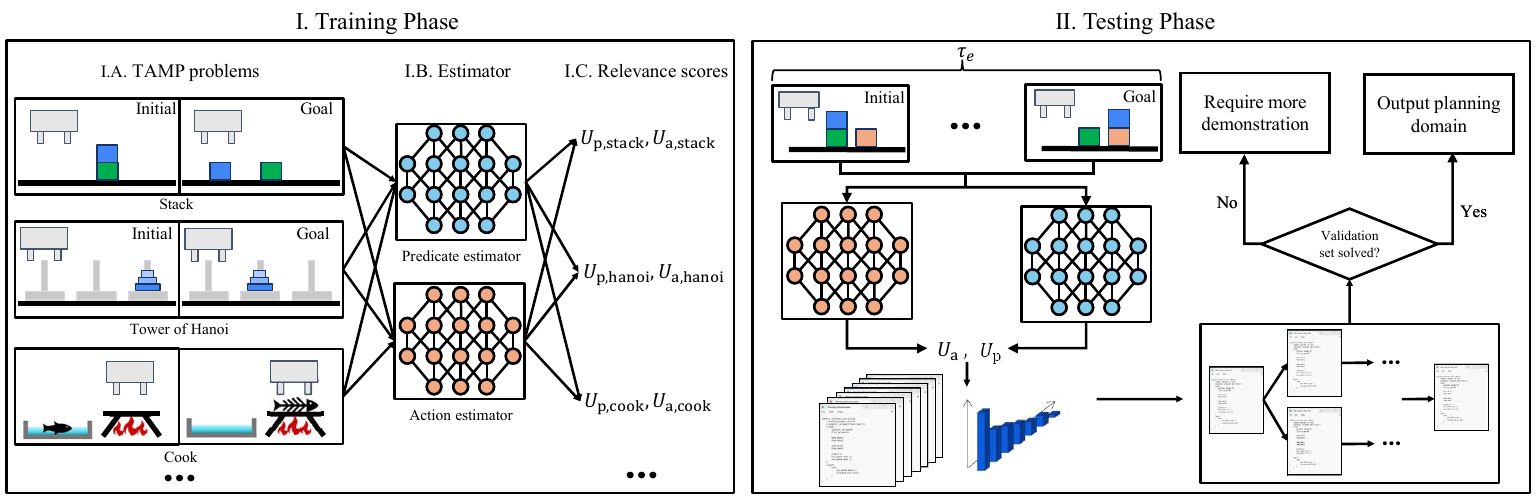}
\vspace{-2mm}
\caption{Our overall framework for domain inference. Left: The training phase of our method. Right: the testing phase of our method.}
\label{fig:Framework}
\vspace{-5mm}
\end{figure*}

% Run on a real robot
% duitribution
\section{Problem Setting}

The standardized planning language used in our work is the Planning Domain Definition Language (PDDL)~\cite{McDermott1998PDDLthePD}. For a given task, the set of objects $\mathbf{O} = \langle o_1, o_2, \dots, o_n \rangle$ includes all relevant objects in the environment, where each object $o \in \mathbf{O}$ is a reference to an entity in the planning problem~\cite{Khodeir2023-xs}. The state function $s: \mathbf{O} \times \mathbf{T} \rightarrow \mathcal{R}^{d \times n}$ provides the continuous state of an object, where $\mathbf{T}$ is the set of time indices at which the state of an object is evaluated. This state is defined by continuous properties, such as pose or temperature.

\vspace{-2.5pt}
\begin{definition}[Full predicate set]
The \textit{full predicate set} $\mathbf{P}$ is a set that encompasses all the predicates required to solve a set of TAMP problems.
\end{definition}
\vspace{-2.5pt}

Each predicate $\mathbf{\mathit{p}} \in \mathbf{P}$ defines a logical property, condition, or relationship among objects (e.g., (\texttt{is\_on}  ?$o_1$ ?$o_2$)) via a classifier function that outputs true or false given the continuous state $ p(s(o_1,t), s(o_2,t), \dots)\in \{\text{true; false}\}$. A ground atom $x$ is a predicate combined with the objects and the classification result (e.g., (\texttt{is\_on} $o_1$ $o_2$)).
The logical state of an environment is a collection of all ground atoms for objects in the environment
$\mathbf{X} = \langle \mathbf{\mathit{x_1}},\mathbf{\mathit{x_2}},\mathbf{\mathit{x_3}},\dots, \mathbf{\mathit{x_n}} \rangle$.

\vspace{-2.5pt}
\begin{definition}[Full action set]
The \textit{full action set} $\mathbf{A} $ is a set of actions that logically describe possible actions the robot may execute for a set of TAMP problems. % defined in \Cref{experiments}.
\end{definition}
\vspace{-2.5pt}

Each logical action $\mathbf{\mathit{a}} \in \mathbf{A}$ has a corresponding precondition $\mathbf{Pre} = \langle\mathbf{\mathit{p_1}},\mathbf{\mathit{p_2}}, ...\rangle$,
which is a set of predicates that must be true to trigger the action, and a postcondition $\mathbf{Eff} = \langle \mathbf{\mathit{p_1}},\mathbf{\mathit{p_2}}, ...\rangle$, which is a set of predicates that indicates the change in the logical environmental state after the action is executed. The transition between logical states before and after the actions is considered to be deterministic, $ \mathbf{X_\mathrm{before}} \times \mathbf{\mathit{a}} \rightarrow \mathbf{X_\mathrm{after}}$. 
A task plan $\mathit{\pi} = \langle \mathbf{\mathit{a_1}},\mathbf{\mathit{a_2}},\mathbf{\mathit{a_3}},..., \mathbf{\mathit{a_n}} \rangle$ is a sequence of logical actions that change the initial logical environmental state to the goal logical environmental state. 
A (logical) trajectory $\mathbf{\tau}$ is the sequence of logical environmental state and logical action at each step during task execution, $\mathbf{\tau} =\langle (\mathbf{X_{1}}, a_{1}), (\mathbf{X_{2}}, a_{2}), \dots, (\mathbf{X_{n}}, a_{n}) \rangle$.

\vspace{-2.5pt}
\begin{definition}[Full motion planner set]
The \textit{full motion planner set} $\mathbf{C}$ is a set of motion planners that produce the corresponding trajectories for actions in $\mathbf{A}$.
\end{definition}
\vspace{-2.5pt}

For each $\mathbf{\mathit{a}}\! \in\! \mathbf{A}$, there is a corresponding $\mathbf{\mathit{c}}\! \in\! \mathbf{C}$ that verifies the feasibility of motion plans for the action and generates the plan if feasible. For example, the motion planner for the action `Pick' will generate a collision-free trajectory for the robot to grasp an object and lift it from the table.

Using the concepts defined above, we can now define a TAMP problem. Any TAMP problem $q$ can be defined by the set of objects, the set of complete predicates, the set of full actions, the initial logical environmental state and the logical environmental state of the goal, $q = \langle\mathbf{O}, \mathbf{P}, \mathbf{A}, \mathbf{X_\mathrm{init}}, \mathbf{X_\mathrm{goal}},\rangle$. A TAMP problem set $\mathbf{Q}$ is a set of unsolved TAMP problems $\mathbf{Q} = \langle q_1,q_2,\dots\rangle$. 

% A planning domain consists of predicates and actions.
The planning domain $\mathbf{D}$ associated with a task is characterized by a set of logical predicates $\mathbf{P_{\mathbf{D}}} \subseteq \mathbf{P}$, a set of logical actions $\mathbf{A_{\mathbf{D}}} \subseteq \mathbf{A}$, and the corresponding motion planner set $\mathbf{C_{\mathbf{D}}} \subseteq \mathbf{C}$. Formally, the planning domain is defined as $\mathbf{D} = \langle\mathbf{P_{\mathbf{D}}}, \mathbf{A_{\mathbf{D}}}, \mathbf{C_{\mathbf{D}}}\rangle$. The domain set $\omega_{\mathbf{D}}$ contains all the names of the predicates in $\mathbf{P_{\mathbf{D}}}$ and actions in $\mathbf{A_{\mathbf{D}}}$. The planning domain $\mathbf{D}$ can be constructed from the domain set $\omega_{\mathbf{D}}$ by designing the preconditions and postconditions for each action and writing the domain into a PDDL file following the PDDL syntax.

A TAMP problem $q$ can be addressed using various planning domains, each with differing performance based on its quality. To evaluate expected performance, we propose two criteria: completeness and optimality, to measure how well a planning domain solves the TAMP problem.

\vspace{-2.5pt}
\begin{definition}[Complete domain]
A complete domain is a domain that includes all predicates and actions needed to solve a TAMP problem.
\end{definition}
%\vspace{-2.5pt}

%\vspace{-2.5pt}
\begin{definition}[Optimal domain]
An optimal domain is a complete domain that contains the least number of predicates and actions needed to solve a TAMP problem.
\end{definition}
\vspace{-2.5pt}

Given $\mathbf{A} $, $\mathbf{P}$, and $\mathbf{C}$, at the testing phase, our system aims at utilizing one (or a few) example trajectory $\mathbf{\tau_e}$ and a validation problem set containing $m$ unsolved TAMP problems, $\mathbf{Q}_v = \langle q_1, q_2, \dots, q_m \rangle$, to produce the optimal planning domain 
$\mathbf{D_\mathrm{optm}} = \langle\mathbf{P_\mathrm{optm}},\mathbf{A_\mathrm{optm}},\mathbf{C_\mathrm{optm}}\rangle$ that can reproduce $\mathbf{\tau_e}$ and generalize to similar TAMP tasks, even with shuffled object poses and varied object numbers. We assume $\mathbf{A} $ and $\mathbf{P}$ are given, as prior studies provide methods for generating predicates and actions~\cite{Kumar2023-nk, Silver2023-mi, Athalye2024-lo, Liang2024-hf}. While these approaches do not learn complete planning domains from scratch, their findings form the basis of this work.

\section{Methodology}

%: (1) The most relevant planning domain, $\mathbf{D_\mathrm{top}}$, is predicted by the predicate and action estimators. (2) The predicted domain, $\mathbf{D_\mathrm{top}}$, is used as the initial hypothesis for domain optimization, a specific instance of combinatorial optimization, where the problem is to find the smallest subset of predicates and actions associated with the example trajectory.
% In this process, we keep adding or removing predicates and actions to the planning domain to solve the validation problem set $\mathbf{Q}_v$. (3) The optimal domain is returned when $\mathbf{Q}_v$ is fully solved; otherwise, more example trajectories from the same type of task are needed.

As shown in \Cref{fig:Framework}, our method proceeds in two phases: a training phase and a testing phase. In the former, we use a training dataset of TAMP problems 
to learn a domain estimator that outputs the relevance of predicates and actions. In the testing phase, we use one or a few demonstration trajectories and a validation problem set to generate an optimal planning domain that does not violate the logical sequences in the demonstration trajectories. 

More concretely, in \Cref{fig:Framework} (I) the training dataset includes information on TAMP problems with their associated planning domains. The predicate and action estimators output a relevance score $u \in [0, 1]$ for each predicate and action, estimating their relevance of being part of the planning domain for a given problem $q$. The scores of predicates (actions) are denoted as $u_\mathrm{p} \in \mathit{U_\mathrm{p}} $ ($u_\mathrm{a} \in \mathit{U_\mathrm{a}} $), where the predicate (action) score set $\mathit{U_\mathrm{p}}$ ( $\mathit{U_\mathrm{a}}$) contains the score for all predicates (actions).

During testing, the inputs are an example trajectory $\mathbf{\tau_e}$ and a validation problem set $\mathbf{Q}_v$. As shown in \Cref{fig:Framework} (II), the testing phase involves two main steps. First, the human demonstration (sensory data or human inputs) is converted into $\mathbf{\tau_e}$ to predict a distribution of relevant planning domains. 
Second, the predicted planning domains are refined by a search algorithm to find the optimal planning domain. If any of the domains explored during the search successfully solve all problems in the validation set $\mathbf{Q}_v$, the system returns the optimal domain $\mathbf{D_\mathrm{optm}}$. Otherwise, additional example trajectories of the same task are required.

\subsection{Predicate and Action Estimators}

\begin{figure}
\centering
\includegraphics[width=.9\linewidth]{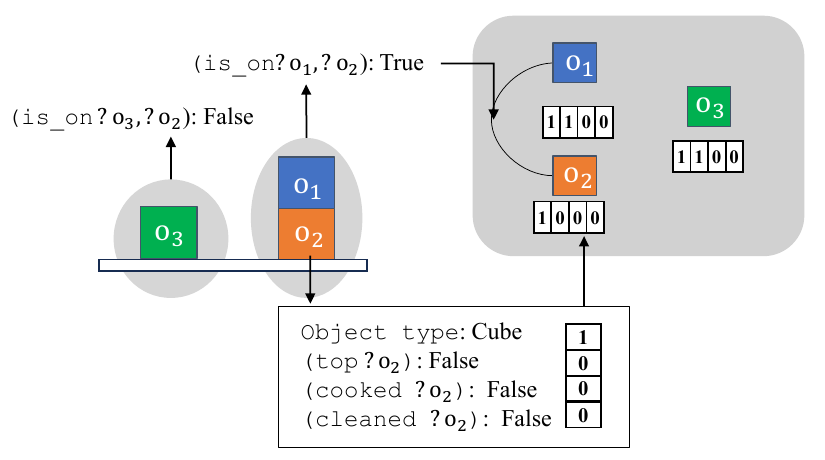}
\vspace{-3mm}
\caption{Example of formulating logical states as a scene graph: (1) Unary predicates: Using the orange cube as an example, object `cube' has an index of 1,  (\texttt{top}, $?o$), (\texttt{cleaned}, $?o$) and (\texttt{cooked}, $?o$) are all false. The feature node vector becomes $[1,0,0,0]$. We repeat this process for the other cubes to obtain the corresponding node features (2) Binary predicates: For example, the predicate (\texttt{is\_on}, $?o_1$, $?o_2$ ) is true for the blue and orange cubes. Thus, an edge is included to connect them. Meanwhile, (\texttt{is\_on}, $?o_3$, $?o_2$ ) is false, so there is no edge between the green and the orange cube.}
\label{fig:typeA}
\vspace{-6mm}
\end{figure}

% The predicates are formulated as node features and edges. One node in the graph represents one object. 
To train the predicate and action estimators, the initial and goal logical states 
$(\mathbf{X_\mathrm{init}},\mathbf{X_\mathrm{goal}})$ of a TAMP problem are represented as scene graphs, with nodes representing objects. As is shown in \Cref{fig:typeA}, unary predicates defining individual object states are encoded as Boolean values in the node feature vector. Binary predicates representing relationships between objects are formulated as edges connecting nodes with one-hot encoded edge features. If predicates involve more than two objects, the objects are connected in pairs, with each connection sharing the same edge feature.

%------------------------------------------------------------------------------
% One example of how the logical states are formulated as graphs (node feature formulation). 

% Overview of the estimator structure: GAT+MLP. The output of estimators are relevance scores. 

Each estimator processes the scene graphs through several graph attention convolution layers (GATConv). The output from the final GATConv layer is passed through a multilayer perceptron (MLP). A sigmoid function is applied at the end to produce an output vector with elements between 0 and 1. Each output vector element indicates the relevance score of a specific predicate $p \in \mathbf{P}$ (action $a \in \mathbf{A}$) with respect to the input scene graph. The relevance score for a domain set is computed in the following manner:

\vspace{-3mm}
\begin{equation}
u_{\mathbf{D}} = \prod_{p \in \omega_{\mathbf{D}}} u_\mathrm{p} \prod_{p \notin \omega_{\mathbf{D}}} (1-u_\mathrm{p}) \prod_{a \in \omega_{\mathbf{D}}} u_\mathrm{a} \prod_{a \notin \omega_{\mathbf{D}}} (1-u_\mathrm{a})
\label{eq1}
\end{equation}

% Explain each term in eq1. 
In \Cref{eq1}, $u_\mathrm{p}$ represents the score for a predicate, while $u_\mathrm{a} $ represents the score for an action. 
The scores for all possible domain sets can be computed using \Cref{eq1}. The domain set with the highest relevance score is denoted as $\omega_{\mathbf{D}}^\mathrm{top}$. The predicates and actions excluded from $\omega_{\mathbf{D}}^\mathrm{top}$ are ranked by their relevance score in a descending manner in a priority list $\mathbf{\mathit{L}}$. Similar score computation schemes can be found in learning-based incremental planning methodologies~\cite{Silver2021-mv}.

\subsection{Dataset Generation}

% To generate the training data, we sample the initial and goal states and label them with relevant predicates and actions.
To train the predicate and action estimator, we require a training dataset of TAMP problems and their associated planning domains. This dataset is generated using a sampling-based approach that randomly samples various planning problems in the simulation. Each TAMP problem is labeled with the relevant predicates and actions for their planning domain, assigning a score of 1 to those crucial for solving the problem and a score of 0 to irrelevant ones. The solvability of TAMP problems are verified via a traditional TAMP planner ~\cite{Khodeir2023-so} deployed on a compute cluster.

% Using unstacking as an example, we formulate initial and goal states as scene graphs: (1) Unary predicates: The dashed blue cube in the goal scene has an object index 1; the cube is on top, so (\texttt{top}, $?o$) is true. The cube is not cleaned or cooked, so both (\texttt{cleaned}, $?o$) and (\texttt{cooked}, $?o$) are false. The resulting node feature vector is $[1,1,0,0]$. (2) Binary predicates: In the initial scene, the predicate (\texttt{is\_on}, $?o_1$, $?o_2$ ) is true for the blue and orange cubes. Thus, a bidirectional edge, represented by a solid line, is formulated to connect them.

\subsection{Precondition and Postcondition Generation} \label{PP}

In the testing phase, we obtain $\omega_{\mathbf{D}}^\mathrm{top}$ from the predicate and action estimators. Then, we must convert $\omega_{\mathbf{D}}^\mathrm{top}$ into an executable PDDL domain $\mathbf{D_\mathrm{top}}$ by determining the preconditions and postconditions for each action. This process depends on finding the commonalities across instances of the same action from the human demonstration.

\vspace{-2.5pt}
\begin{definition}[Pre-image]
The logical world state for the last time step before the action is triggered.
\end{definition}
\vspace{-3pt}

\vspace{-2.5pt}
\begin{definition}[Post-image]
The logical world state for the first time step after the action is completed.
\end{definition}
\vspace{-3pt}

The precondition is determined by identifying the intersections of the pre-images of the same action across multiple instances. We first extract the pre-images for all actions and then group them by the action type. Next, we find the intersection of all pre-images for the same action to be the precondition.
Similarly, for generating postconditions, we start by extracting the post-images for all actions and grouping them. The intersection of the post-images for each action is then calculated. This intersection is compared to the corresponding precondition, and any unchanged predicates are removed. The resulting set of predicates is returned as the postcondition.
% Collecting all actions, the planning domain is automatically written into a PDDL file.

\begin{figure*}[htbp]
    \centering
    \includegraphics[width=0.96\textwidth]{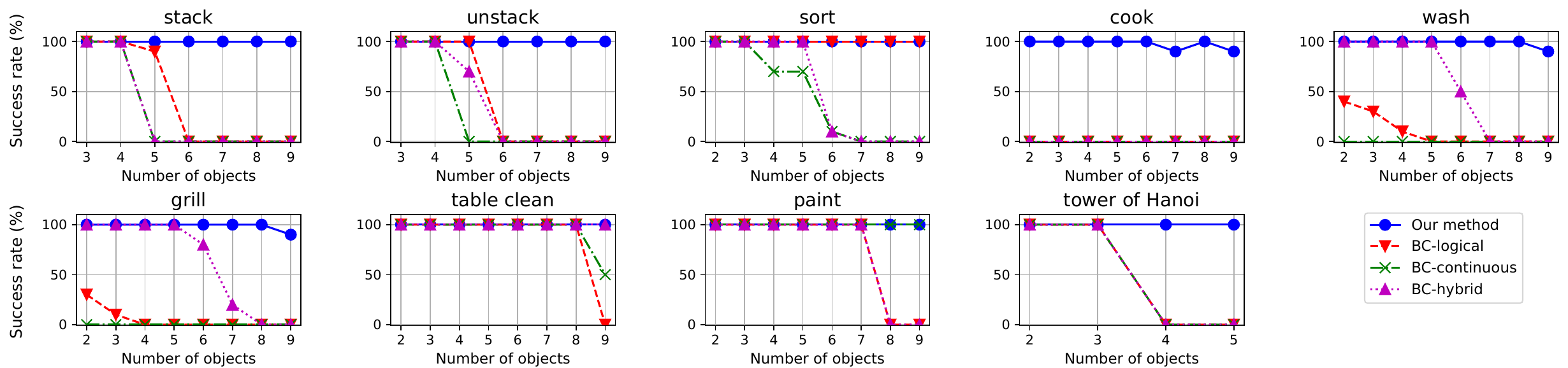} % Adjust width accordingly
    \vspace{-2.5mm}
    \caption{Success rates of basic planning tasks with increasing objects, compared to baselines (one demonstration per task).}
    \label{fig:planning_success}
    \vspace{-1.5mm}
\end{figure*}

\begin{figure*}[htbp]
    \centering
    \includegraphics[width=0.95\textwidth]{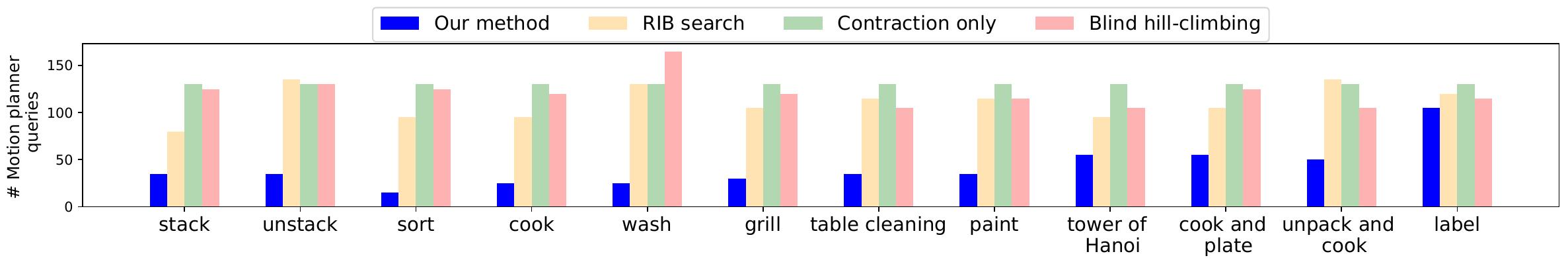}  % Adjust width accordingly
    \vspace{-2.5mm}
    \caption{Comparison of motion planner queries for domain optimization, ensuring all methods find the optimal domain.}
    \label{fig:search_cost}
    \vspace{-5mm}
\end{figure*}

\subsection{Domain Optimization by Generate-and-Test Search} 
\label{DP}

The domain optimization process is based on the generate-and-test search method, a heuristic search technique that involves backtracking~\cite{Mahmood2013-ye}. Starting from $\mathbf{D_\mathrm{top}}$, the optimization process iterates through all possible domains until the domain successfully solves the validation problem set $\mathbf{Q}_v$ with the minimal number of predicates and actions.
There are two assumptions made for the search:

\newtheorem{assumption}{Assumption}
\vspace{-2.5pt}
\begin{assumption}
    If a domain $\mathbf{D}$ is incomplete, then any subdomain $\mathbf{D_s} \subseteq \mathbf{D}$ is incomplete.
\end{assumption}
\vspace{-3pt}

\vspace{-3pt}
\begin{assumption}
    If a domain $\mathbf{D}$ is complete, then the optimal domain $\mathbf{D_\mathrm{optm}} \subseteq \mathbf{D}$.
\end{assumption}
\vspace{-3mm}

\begin{algorithm}
\caption{Domain Optimization Algorithm.}
\begin{algorithmic}[1]

\State $\mathbf{D_\mathrm{perturb}} = \mathbf{D_\mathrm{top}}$ 
\Comment{Initialization.}
\State $is\_solved =plan(\mathbf{D_\mathrm{perturb}},\mathbf{Q}_v)$
\State $element\_list = \omega_\mathrm{\mathbf{D}}^\mathrm{top}$

\State \Comment{$top(\mathbf{\mathit{L}}$) returns the highest-scored element from $\mathbf{\mathit{L}}$.}

\While{not $is\_solved$}
    \State $\mathbf{D_\mathrm{perturb}} = add(\mathbf{D_\mathrm{perturb}},top(\mathbf{\mathit{L}}))$ 
    \State $is\_solved =plan(\mathbf{D_\mathrm{perturb}},\mathbf{Q}_v)$
    \State $element\_list.insert(top(\mathbf{\mathit{L}}))$
    \State $\mathbf{\mathit{L}}.pop( top(\mathbf{\mathit{L}}))$
\EndWhile 

\State $\mathbf{D_\mathrm{optm}} = \mathbf{D_\mathrm{expanded}} = \mathbf{D_\mathrm{perturb}}$

\For{$ e_i \in element\_list $}
    \State $\mathbf{D_\mathrm{perturb}} = remove(\mathbf{D_\mathrm{optm}},e_i)$
    \State $is\_solved =plan(\mathbf{D_\mathrm{perturb}},\mathbf{Q}_v)$
    \If{$is\_solved$}
        \State $\mathbf{D_\mathrm{optm}} = \mathbf{D_\mathrm{perturb}}$
    \EndIf
\EndFor

\end{algorithmic}
\label{a1}
\end{algorithm}
\vspace{-2mm}

The optimization process aims to add or remove predicates (actions) from the initial guess until the optimal domain is found. The domain $\mathbf{D_\mathrm{perturb}}$ is initialized as $\mathbf{D_\mathrm{top}}$. There are two major stages: expansion and contraction. 

%Step 1: we expand the domain by including more predicates/actions until the validation problems are fully solved; this domain is the temporary d_optm. 
In the expansion stage, $\mathbf{D_\mathrm{perturb}}$ is expanded until it solves the validation problem set $\mathbf{Q}_v$. In each iteration, we move the top predicate(action) from priority list $\mathbf{\mathit{L}}$ and add it to $\mathbf{D_\mathrm{perturb}}$. The algorithm plans with $\mathbf{D_\mathrm{perturb}}$ to solve $\mathbf{Q}_v$. We continue expansion until $\mathbf{D_\mathrm{perturb}}$ can fully solve $\mathbf{Q}_v$.

% Step 2: We remove each element d_optm once. If the domain can still solve validation problems without the removed element, the removed element is redundant. Thus it is permanently removed. If the domain cannot solve validation problems without the removed element, the removed element is critical. Then the element is added back to d_optm.
In the contraction stage, we remove redundant elements from $\mathbf{D_\mathrm{optm}}$ until no more elements can be removed while maintaining domain completeness. The output domain from the expansion stage, $\mathbf{D_\mathrm{expanded}}$, is initialized as $\mathbf{D_\mathrm{optm}}$. In each iteration $i$, we remove one element (either a predicate or an action), $e_i \in \mathbf{D_\mathrm{optm}}$, to produce a perturbed domain $\mathbf{D_\mathrm{perturb}}$. If $\mathbf{D_\mathrm{perturb}}$ fully solves $\mathbf{Q}_v$, the removed element $e_i$ is redundant. Then $\mathbf{D_\mathrm{perturb}}$ becomes the new optimal domain $\mathbf{D_\mathrm{optm}}$. If $\mathbf{D_\mathrm{perturb}}$ fails to solve $\mathbf{Q}_v$, the removed element $e_i$ is critical. In this case, we backtrack to the last $\mathbf{D_\mathrm{optm}}$. This iteration continues until each element in $\mathbf{D_\mathrm{expanded}}$ has been evaluated. 

\section{Experiments}
In this section, we introduce the baselines and experimental tasks. We aim to answer the following questions: (Q1) How effective is planning using the inferred planning domain? (Q2) Can we generalize to more complex and unseen tasks? (Q3) During testing, how much does our method reduce computational costs of domain generation?

\subsection{Planning Baselines}

We compared the planning success rate of our method to behaviour cloning baselines to highlight its improved efficiency and generalizability. All baselines are trained per task type. The graph attention network training is implemented in PyTorch~\cite{paszke2017automatic} and PyTorch Geometric~\cite{Fey2019-lv}.

\textbf{BC-logical}:  Inspired by prior work~\cite{Zhu2021-sx}, this baseline uses logical scene graphs to train a GAT task planning policy. The scene graph formulation is the same as that of our estimator. Using the current and goal logical scene graph as input, BC-logical outputs the next logical action to be executed.

\textbf{BC-continuous}: This baseline is a modification of BC-logical. However, the node features of the input scene graph nodes contain only continuous object states.

\textbf{BC-hybrid}: Inspired by recent papers combining continuous and logical data for imitation learning in TAMP~\cite{Khodeir2023-so, Lin2022-tc}, BC-hybrid deviates from BC-logical by using both continuous and logical states of objects in the node features.

\subsection{Domain Optimization Baselines}

We compare our domain optimization method to several baselines to assess the potential reduction in computational cost for easier tasks, as we learns planning domains for simple tasks that can be used in complex ones.

\textbf{RIB search}: The random-initial-and-blind (RIB) search method randomly samples an initial planning domain and searches through all possible planning domains blindly. This method terminates when the planning domain is complete and no predicates or actions can be further removed.

\textbf{Contraction}: This method deviates from the RIB search method as it starts with an initial planning domain that contains all predicates and actions in $\mathbf{P}$ and $\mathbf{A}$. 

\textbf{Blind hill-climbing}: Inspired by the recent work in automatic planning domain generation~\cite{Kumar2023-nk,Silver2023-mi}, this search method starts with an empty initial planning domain and searches through all possible planning domains via hill-climbing. This method searches for the planning domain with the highest planning success rate while maintaining the minimal planning domain size.

\subsection{Experimental Tasks}

Experiments are developed on the PyBullet simulation~\cite{benelot2018}. The task planner is based on the PDDLStream library~\cite{Garrett2020-cr}, and the motion planner is based on the pybullet-planning library~\cite{Garrett2015-wc}. We designed nine basic tasks and three composed tasks to evaluate generalizability to unseen environments, where the tasks can be described by the same action and predicate set, with varying object counts and positions.

\textbf{Basic tasks}: In \textit{stacking}, the robot needs to stack cubes into a tower in a specific order. In the \textit{unstacking} task, the objective is to disassemble the tower of cubes. In \textit{sorting}, the robot needs to cluster cubes into two groups at different locations. In \textit{washing}, the robot needs to wash food ingredients in the kitchen sink. In \textit{grilling}, the robot needs to grill food on the stove. The \textit{cooking} task involves a combination of washing and grilling food. The \textit{table cleaning} task indicates collecting objects from the table and storing them in a bin. The \textit{painting} task involves drawing random figures on a piece of paper. The last basic task is to solve \textit{Towers of Hanoi}.

\textbf{Composed tasks}: Composed tasks are formed by combining multiple basic tasks. The \textit{unpack-and-cook} task involves unstacking raw materials and cooking them. In \textit{cook-and-plate} task, the robot needs to cook the raw materials and stack them together. The \textit{labeling} task requires the robot to unstack piles of objects and then label each item with a pen.

\subsection{Training and Testing Dataset} 

The training dataset includes 30 demonstrations per basic task, each involving a solvable TAMP problem containing two to four objects. 
The solution trajectory and the human demonstration are provided together as state-action pairs for baseline training.
The test dataset includes 870 different TAMP problems, encompassing basic and composed tasks, with object counts in the planning environment ranging from two to nine. All planning problems are generated by uniformly sampling random object poses across the entire workspace (table). For each object count, 10 test problems are generated with randomized object positions, and each problem must be solved within 30 seconds.

\begin{figure}[t!]
\centering
\includegraphics[width=1\linewidth]{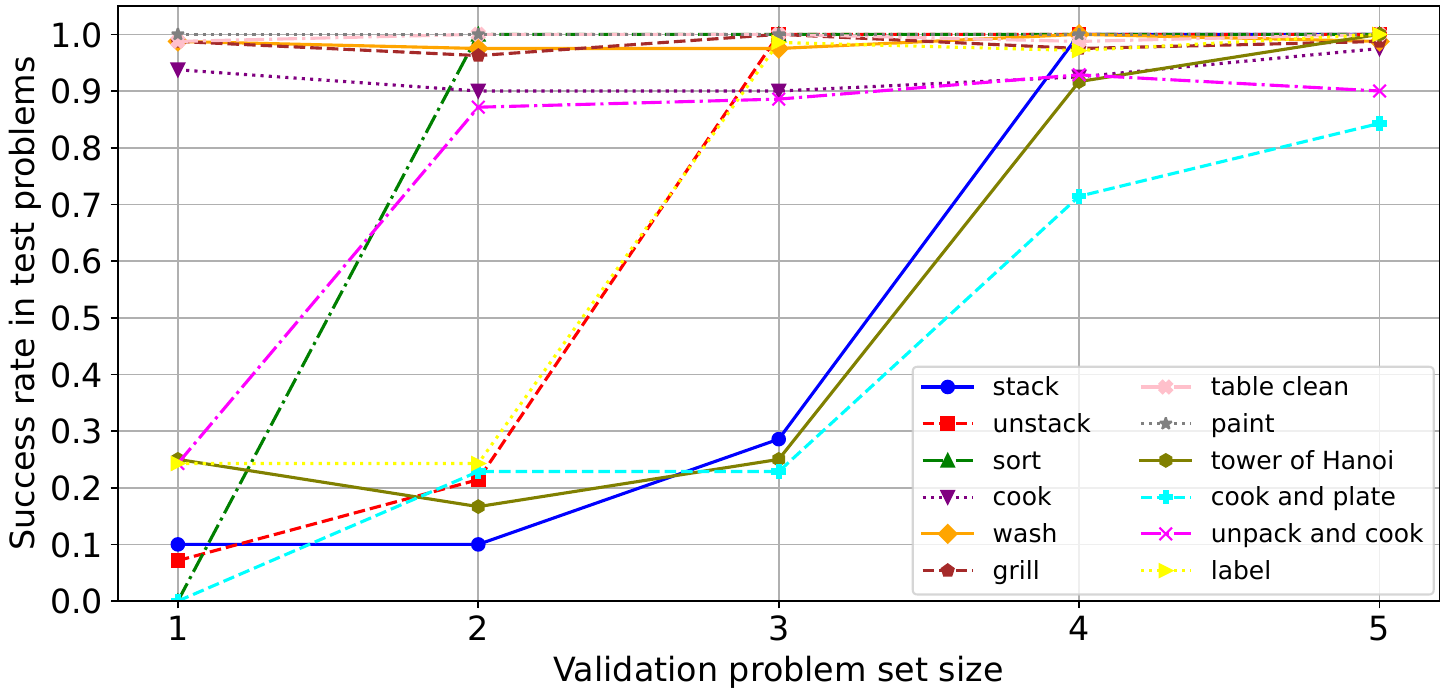}
\vspace{-6mm}
\caption{Planning success rate of the domain inferred with different sizes of validation problem set $\mathbf{Q}_v$ that assist domain optimization in \Cref{a1}.}
\label{fig:validation}
\vspace{-5mm}
\end{figure}

\subsection{Real Robot Experiments}

To demonstrate the effectiveness of our system with real robots, we implemented two real-robot experiments: one involving the Tower of Hanoi and the other focusing on stacking multiple cube towers. We adopt NVIDIA's FoundationPose \cite{Wen_2024_CVPR} to capture visual demonstrations, which provide the 6D pose trajectories of all objects. These poses are then consumed downstream to compute predicates concerning the objects in the scene. The real-robot validation problems are set up manually, with photos of the initial and goal configurations processed in the same way. The inferred planning domain is combined with the ROS Moveit motion planner and deployed on a Franka Research 3 robot.

\begin{figure}[b!]
\vspace{-4mm}
\centering
\includegraphics[width=1\linewidth]{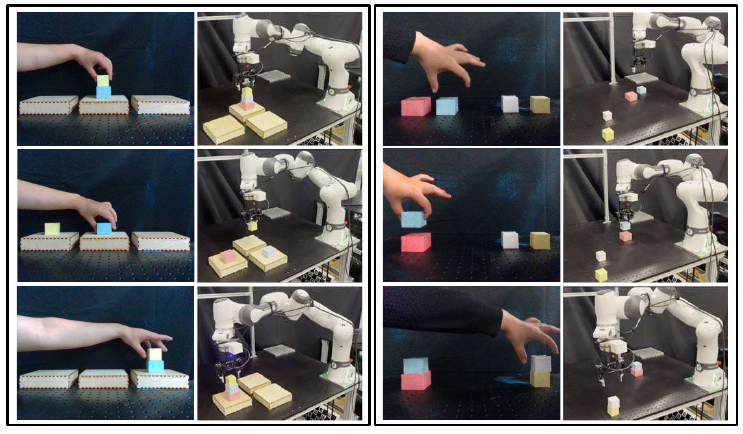}
\vspace{-6mm}
\caption{Real robot experiment. Left: Towers of Hanoi. Right: Stack towers}
\label{fig:realrobot}
\end{figure}

\section{Results and Discussion}

To address (Q1), we evaluate our method on various tasks and compare it to a few baselines. As illustrated in \Cref{fig:planning_success}, our method consistently outperforms the baselines in planning success rate across all basic tasks. While both our method and the baselines perform well on simpler tasks (painting, table cleaning, and sorting), the baselines degrade significantly on more complex tasks with larger decision spaces. Our baseline evaluation aligns with prior studies, reinforcing that behaviour cloning requires more comprehensive training data for generalization \cite{Kumar2023-nk, Silver2023-mi}.

To address (Q2), we evaluate our methods with planning problems that involve an increased number of objects, as well as unseen composed tasks. \Cref{fig:planning_success} shows that our method maintains success rates above 90\% even with up to nine objects, while the behaviour cloning baseline method experiences a sharp decline as the number of objects increases. These results indicate that our method has a significant advantage in generalizing to a more complex experiment setup. Furthermore, as shown in \Cref{tab:success_rates}, our method demonstrates excellent planning performance on generalizing to unseen composed tasks. With just one human demonstration per task, our method successfully infers the planning domain, enabling efficient planning even when the number of objects increases and their positions are shuffled. Since we only have one trajectory per composed task, we did not compare our method with the baselines. Training a behaviour cloning policy with just one trajectory would be trivial.

To answer (Q3), we compare the computational cost of domain optimization to the baselines. The computational cost is assessed based on the number of queries made to the motion planner, which is the most time-intensive component of the optimization process. This metric is chosen because it provides a consistent comparison independent of the computer's performance. As shown in \Cref{fig:search_cost}, our method drastically lowers the computational cost in the optimization process.
The reduction in computational cost is more apparent for basic task types and attenuates when dealing with composed task types because the most relevant domain prediction becomes less accurate. \Cref{fig:validation} shows our method’s testing planning success rate of the domain inferred with different sizes of the validation problem set $\mathbf{Q}_v$ via \Cref{a1}. The results indicate that our method requires only a small $\mathbf{Q}_v$. For many tasks, the correct planning domain can be inferred with just one or two validation problems. Even in more challenging scenarios, five validation problems are sufficient.

\begin{table}[t]
    \centering
    \caption{Success rates for composed planning tasks.}
    \vspace{-2mm}
    \resizebox{0.5\textwidth}{!}{ % Adjust the table size as needed
        \begin{tabular}{lc|ccccccc}
            \toprule
            \multirow{2}{*}{\textbf{Task Type}} & \multicolumn{8}{c}{\textbf{Number of Objects}} \\
            \cmidrule(lr){2-9}
            & & \textbf{3} & \textbf{4} & \textbf{5} & \textbf{6} & \textbf{7} & \textbf{8} & \textbf{9} \\
            \midrule
            Cook and Plate & & 100 & 100 & 100 & 100 & 100 & 40 & 50 \\
            Unpack and Cook & & 100 & 100 & 100 & 100 & 100 & 80 & 50 \\
            Label & & 100 & 100 & 100 & 100 & 100 & 100 & 100 \\
            \bottomrule
        \end{tabular}
    }
    \label{tab:success_rates}
\vspace{-5mm}
\end{table}

The real robot experiment, presented in \Cref{fig:realrobot}, demonstrates the effectiveness of our method on a real-world system. Remarkably, we only need to demonstrate once for the real robot to learn and solve similar TAMP problems.

\section{Conclusion}

This paper proposes a method for automatically inferring planning domains in task and motion planning by integrating deep learning and search. Our approach was evaluated in various planning environments, showing improved planning performance and generalizability compared to existing learning-based TAMP solvers. We also demonstrate a significant reduction in computational costs during domain optimization compared to common search algorithms. Real robot experiments are performed to validate the feasibility of deploying our method in a real-world implementation.

\renewcommand*{\bibfont}{\normalfont\footnotesize}
\balance
\printbibliography
\end{document}